\documentclass[runningheads]{llncs}
\usepackage[T1]{fontenc}
\usepackage{graphicx}
\usepackage{subcaption}
\usepackage{adjustbox}
\usepackage{float}
\usepackage{placeins}
\usepackage{url}

\begin{document}
\title{Automatic Aorta Segmentation with Heavily Augmented, High-Resolution 3-D ResUNet: Contribution to the SEG.A Challenge}
\titlerunning{Automatic Aorta Segmentation: Contribution to the SEG.A Challenge}
%
\author{Marek Wodzinski\inst{1, 2}\orcidID{0000-0002-8076-6246} \and
Henning M\"{u}ller\inst{1, 3}\orcidID{0000-0001-6800-9878}}
\authorrunning{M. Wodzinski and Henning M\"{u}ller}
%
\institute{
$^{1}$University of Applied Sciences Western Switzerland (HES-SO Valais) \\ Information Systems Institute, Sierre, Switzerland \\
$^{2}$AGH University of Krakow, Department of Measurement and Electronics \\ Krakow, Poland \\
$^{3}$ University of Geneva, Medical Faculty, Geneva, Switzerland \\
\email{
wodzinski@agh.edu.pl \\
}
}
\maketitle              
\begin{abstract}
Automatic aorta segmentation from 3-D medical volumes is an important yet difficult task. Several factors make the problem challenging, e.g. the possibility of aortic dissection or the difficulty with segmenting and annotating the small branches. This work presents a contribution by the MedGIFT team to the \textit{SEG.A} challenge organized during the MICCAI 2023 conference. We propose a fully automated algorithm based on deep encoder-decoder architecture. The main assumption behind our work is that data preprocessing and augmentation are much more important than the deep architecture, especially in low data regimes. Therefore, the solution is based on a variant of traditional convolutional U-Net. The proposed solution achieved a Dice score above 0.9 for all testing cases with the highest stability among all participants. The method scored 1st, 4th, and 3rd in terms of the clinical evaluation, quantiative results, and volumetric meshing quality, respectively. We freely release the source code, pretrained model, and provide access to the algorithm on the Grand-Challenge platform.

\end{abstract}
\section{Introduction}

\subsection{Overview}

The automatic segmentation of the aorta and its branches is one of the most important tasks for cardiovascular system diagnosis and interventions~\cite{aorta_review}. Aorta diseases like stenosis or dissections pose significant threat to patient's health. Early, automatic detection of such diseases is crucial to ensure timely diagnosis and treatment~\cite{tortora,nienaber}. The problem of aorta segmentation seems to be relatively straightforward for the novel deep learning solutions~\cite{survey1,survey2} since aorta is clearly visible structure in the magnetic resonance images (MRI), computed tomography angiography (CTA) or sometimes even in computed tomography (CT) without contrast agents. Nevertheless, several challenges make the segmentation task demanding: (i) the possibility of aortic dissections, stenosis, and other morphological changes that increase the data heterogeneity, (ii) the importance of small branches, (iii) the difficulty in acquiring high-quality annotations, (iv) differences in equipment and acquisition protocols between medical centers. These problems, among others, motivated researchers to organize the SEG.A Challenge during MICCAI 2023 conference~\cite{challenge}. This work presents the contribution of the MedGIFT team to the challenge. 

\subsection{Related Work}

The manual aorta segmentation is a time-consuming task prone to human errors. Annotating even a single high-resolution case takes several hours~\cite{hahn}, thus it is crucial to speed-up the process by automatic or semi-automatic algorithms. There are only a few studies focused on the fully automatic segmentation of aorta and the whole vessel tree~\cite{shahzad,chen}. The methods for the automatic aorta segmentation can be divided into different types: (i) deformable models, (ii) tracking models, (iii) deep learning-based models, (iv) others~\cite{data2}. In this work we focus only on the learning-based methods, for a detailed description of other techniques we refer to the most recent survey~\cite{data2}. Majority of the existing contributions use some variant of the UNet architecture~\cite{unet,trullo,lopez,cao,fantazzini,howard,berhane,hahn,chen,yu,jin,zhong,sieren}. Almost all of the works use the deep network directly on input volume, after downsampling that makes it possible to train the architecture with GPU acceleration. In contrast, the work by Fantazzini~\textit{et al.}~\cite{fantazzini} employs a combined 2-D / 3-D approach where the 3-D network performs the initial aorta segmentation and then the 2-D networks fine-tune the segmentation mask to capture fine-details. The reported Dice coefficient (DSC) varies from 0.82~\cite{lopez} to even 0.97~\cite{chen}, however they cannot be directly compared because different datasets, splits, and even modalities are used. The reported dataset size varies from just several cases~\cite{lopez} to more than thousand~\cite{hahn}. The reported processing time also varies strongly from a fraction of second~\cite{berhane} or even several minutes~\cite{hahn,yu} per case. Since the currently available contributions cannot be directly compared, the SEG.A challenge is strongly motivated and may provide a large-scale benchmark of the current-state-of-the-art. In this work, we combine the strongest aspects of the already existing contributions and propose high-quality, stable algorithm for the automatic segmentation of the aortic vessel tree.

\subsection{Design Choices}

Several factors motivated our design choices:
\begin{enumerate}
    \item The ground-truth annotation of the aortic vessel tree is time-consuming. The provided training set for the method development consists of only 56 cases. Therefore, the proposed solution should operate in an extremely low data regime.
    \item One of the most important factors is the algorithm stability and robustness. Therefore, it is crucial to ensure that the solution works correctly even in the most difficult scenarios.
    \item The training set consists of CTA scans from various centers with different intensity distributions, thus, requiring a careful approach to the intensity normalization and training augmentation.
    \item The small aorta branches are hardly visible and substantial downsampling may reduce their segmentation quality.
\end{enumerate}

These factors motivated us to propose a traditional 3-D CNN based on the UNet architecture, instead of using more novel architectures based on e.g. Vision Transformers~\cite{transformers}. We argue that with such a low amount of training data the network architecture is significantly less important than the data preparation, preprocessing, and augmentation. Moreover, it may even turn out that inductive bias introduced by CNNs is beneficial. Therefore, we decided to:
\begin{enumerate}
    \item Use 3-D residual UNet with proven usefulness in numerous tasks.
    \item Apply heavy data augmentation consisting of several randomly permuted transformations providing substantial variability to the training set.
    \item Use relatively large volume shape: 400x400x400 requiring significant computational resources and training time.
\end{enumerate}

\subsection{Contribution}

In this work, we present our contribution to the SEG.A challenge organized during the MICCAI 2023 conference. We propose and evaluate 3-D CNN architecture based on residual UNet and show the importance of data preprocessing and augmentation. The proposed contribution is among the best-performing ones and arguably the most stable one in terms of the Dice coefficient. We freely release the source code, pretrained model, and provide access to the algorithm on the Grand-Challenge platform.

\section{Method}

\subsection{Preprocessing}

The input volumes are resampled to 400x400x400, clipped to the [-700, 2300] range, and then normalized to the [0-1] range before any further operations.

\subsection{Augmentation}

During training, the data is augmented by: (i) random affine transformation, (ii) random intensity transformation, (iii) random Gaussian noise, (iv) random flipping (all axes), (v) random motion artifacts, (vi) random anisotropy transformation, (vii) random Gaussian blurring. The transformations are applied in random order, each with a probability equal to 0.5. The augmentation is implemented using TorchIO library~\cite{torchio}. Moreover, before training the data was offline augmented by elastic transformations to generate 15000 augmented cases. The reason for performing the elastic augmentation before training was related to the computational complexity of this transformation and the resulting CPU bottleneck.  

\subsection{Deep Network}

A dedicated neural network based on the 3-D ResUNet architecture was implemented. The network takes as input Bx1xHxWxD volumes and outputs segmentations with the same shape. The details related to the network architecture are available in the associated repository~\cite{source_code}.

\subsection{Objective function \& Training}

The objective function is a linear combination of two loss terms: (i) Soft Dice Loss, and (ii) Focal Loss, with the same weight for both the loss terms~\cite{monai}.

Fully supervised approach was employed for the network training. The AdamW was used as the optimizer, with an exponentially decaying learning rate scheduler. The network was trained until convergence on the validation set. Additional experiments with 5-fold cross-validation were performed. No ensembles were employed, the best final model was used for the final Docker container.

\subsection{Inference \& Postprocessing}

The inference consists of the following steps:
\begin{enumerate}
    \item Loading a given case.
    \item Resampling the input case to 400x400x400, clipping the intensity, and normalizing it to the [0-1] range.
    \item Calculating the prediction (without thresholding).
    \item Resampling the prediction by linear interpolation to the original shape and thresholding.
    \item Performing the postprocessing based on the connected component analysis to leave only the largest connected component (for further meshing only).
    \item Saving the calculated segmentation mask.
\end{enumerate}

\subsection{Meshing}

The calculated segmentation mask is meshed using the Discrete Marching Cubes available in the VTK library. Afterward, the mesh is smoothed by WindowedSincPolyDataFilter with the following parameters for the surface meshing: (i) boundary smoothing set: false, (ii) feature edge smoothing set: false, (iii) number of iterations: 25, (iv) feature angle: 120, (v) pass band: 0.001, (vi) non-manifold smoothing: true. The following arguments are used for the mesh that was further used for volumetric meshing: (i) boundary smoothing set: true, (ii) feature edge smoothing set: true, (iii) number of iterations: 30, (iv) feature angle: 120, (v) pass band: 0.001, (vi) non-manifold smoothing: true. Moreover, all holes in the mesh are closed to make it watertight. Small branches are slightly extended by morphological operations (before meshing) to ensure that the TetGen can create the volumetric mesh successfully.

\subsection{Dataset}

\begin{figure*}[!htb]
	\centering
    \includegraphics[scale=0.72]{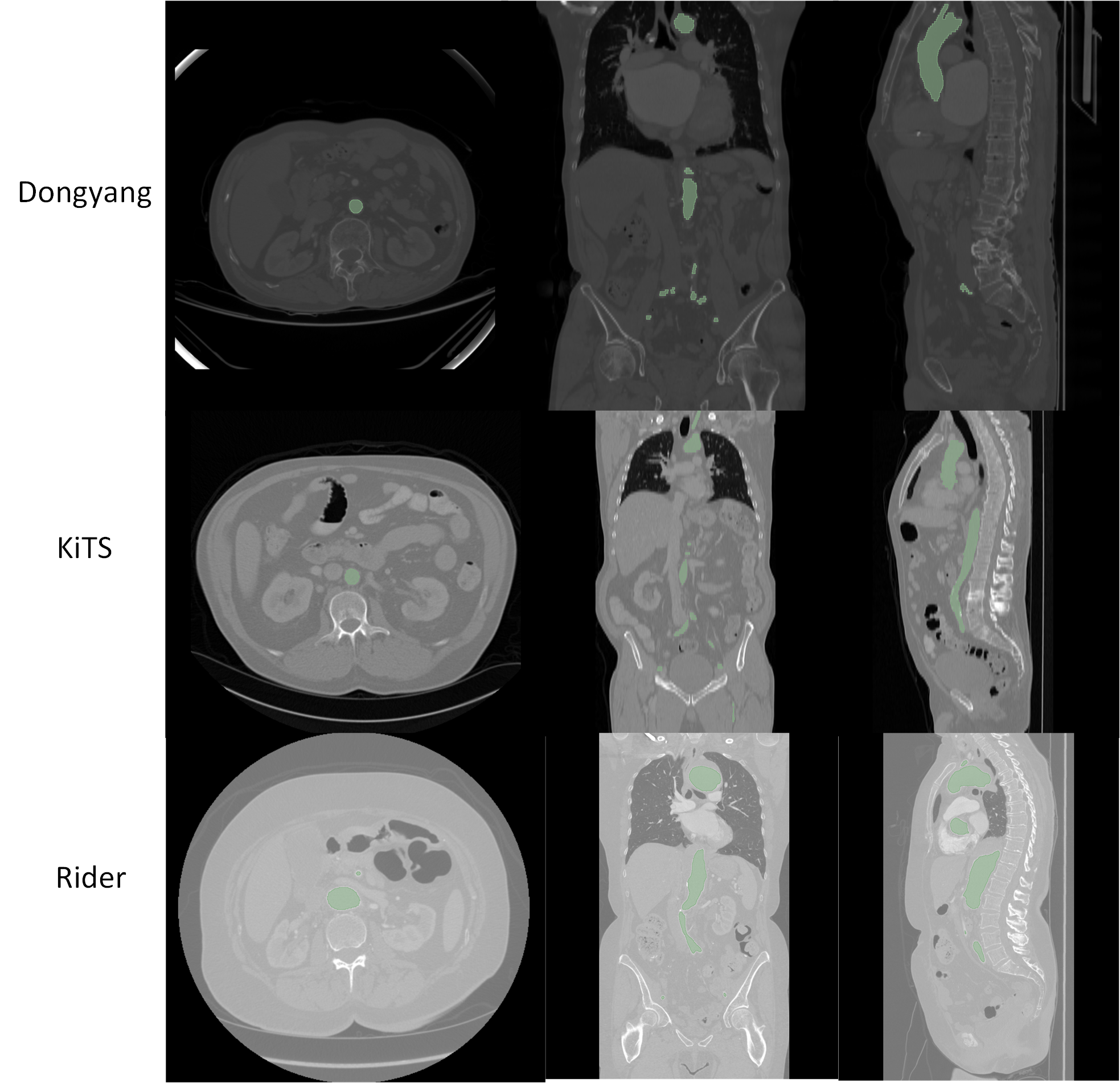}
    \caption{Exemplary cases from each medical center. Please note the differences related to the intensity distributions (unrelated to the visualization window), field of view, and spatial resolution (e.g. Rider cases have significantly smaller voxel size).}   
    \label{fig:dataset}
\end{figure*}

The training data consists of 56 annotated cases provided by the challenge organizers. The scan contains the aorta and all its branches acquired with the computed tomography angiography (CTA)~\cite{data1,data2,aorta_review}. The dataset, unlike other publicly available datasets, annotates the whole aortic vessel tree: the ascending aorta and the branches into the thoracic aorta, abdominal aorta, head/neck area, and iliac arteries branching to the legs. The cases were annotated using three datasets: KiTS~\cite{kits1,kits2}, RIDER~\cite{rider2}, and Dongyang Hospital. Several of the cases exhibit pathologies, such as aortic dissections or abdominal aortic aneurysms. The exemplary scans from each medical center are presented in Figure~\ref{fig:dataset}. Only the data provided by the challenge organizers was used during training. No external data was used. No pretrained networks were used.

\subsection{Experimental Setup}

Apart from the final submission to the Grand Challenge platform, several ablation studies were performed to verify: (i) the influence of resampling to a lower resolution, (ii) the influence of cost function, and (iii) the impact of data augmentation. All these ablation studies were performed using a 5-fold cross-validation. All the performed experiments were trained until convergence. The final challenge submission comes from an experiment with the following training hyper-parameters:
\begin{itemize}
    \item Initial learning rate: 0.001
    \item Iteration size: 64
    \item Batch size: 16
    \item Decay rate: 0.999
    \item Cost function: Dice-Focal Loss
    \item Max gradient value: 2
    \item Max Gradient norm: 10
    \item Optimizer: AdamW
    \item Optimizer weight decay: 0.005
    \item Volume resolution: 400x400x400
    \item Augmentation: Elastic, Affine, Contrast, Gaussian Noise, Flipping, Motion Artifacts, Anisotropy.
\end{itemize}

\subsection{Source Code}

All network, training, augmentation, and inference hyperparameters are reported in the repository. The source code of the proposed contribution is available at~\cite{source_code}. The pretrained model is available upon request, as well as the access to the Grand Challenge algorithm~\cite{algorithm}.

\section{Results}

\subsection{Aorta Segmentation}

\begin{figure*}[!htb]
	\centering
    \includegraphics[scale=0.65]{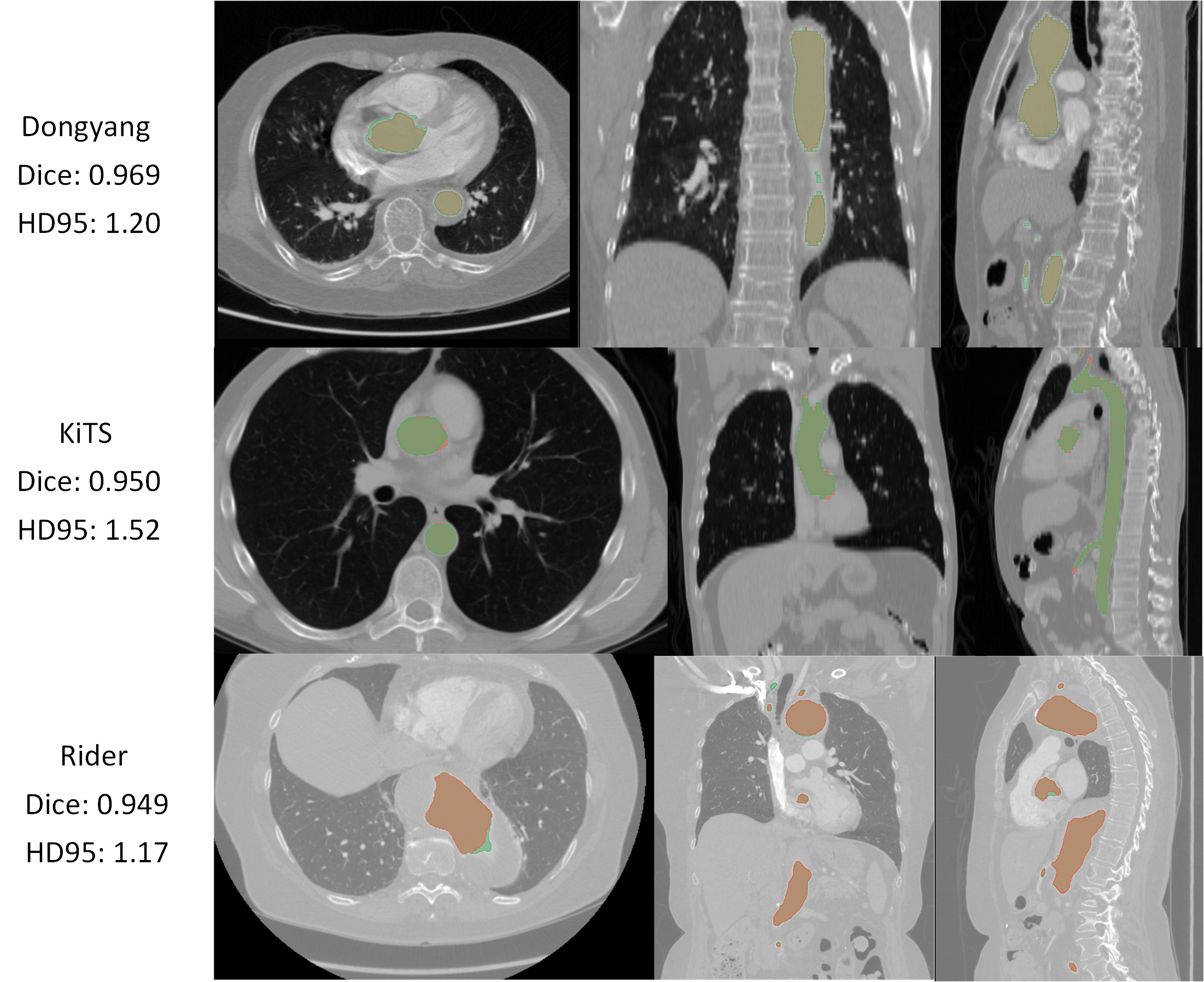}
    \caption{Exemplary visualizations from the internal validation set for cases from each data source. The ground-truths are shown in green and the calculated segmentation masks in red.}   
    \label{fig:results}
\end{figure*}

The results of aorta segmentation are evaluated using the Dice coefficient and the 95th percentile of the Hausdorff distance. The quantitative results for the 5-fold cross-validation are reported in Table~\ref{tab:results} and qualitative visualizations are presented in Figure~\ref{fig:results}. The results for the external test set are reported in Table~\ref{tab:results_test} and compared to other challenge participants.

\begin{table*}[!htb]
\centering
\caption{The quantitative results on a 5-fold cross-validation (using the 56 provided training cases).}
\renewcommand{\arraystretch}{1.0}
\footnotesize
\resizebox{0.9\textwidth}{!}{%
\begin{tabular}{lcc}
\label{tab:results}
Method & Avg. DSC $\uparrow$ & Avg. HD95 [mm] $\downarrow$ \tabularnewline
\hline
\multicolumn{3}{c}{Ablation Studies - Resolution (Dice + Focal, full augmentation)} \tabularnewline
\hline
$400^3$ (proposed) & 0.9441 & 1.89 \tabularnewline
$256^3$ & 0.9235 & 2.45 \tabularnewline
$224^3$ & 0.9182 & 2.82 \tabularnewline
$160^3$ & 0.9081 & 3.56 \tabularnewline
\hline
\multicolumn{3}{c}{Ablation Studies - Cost Function ($400^3$, full augmentation)} \tabularnewline
\hline
Dice + Focal (proposed) & 0.9441 & 1.89 \tabularnewline
Dice & 0.9362 & 2.24 \tabularnewline
Dice + Cross-Entropy & 0.9215 & 2.87 \tabularnewline
Focal & 0.8748 & 5.59 \tabularnewline
\hline
\multicolumn{3}{c}{Ablation Studies - Data Augmentation ($400^3$, Dice + Focal)} \tabularnewline
\hline
Full Augmentation (proposed) & 0.9441 & 1.89 \tabularnewline
Without Elastic & 0.9331 & 2.37 \tabularnewline 
Without Geometric & 0.8941 & 3.42 \tabularnewline 
Without Intensity & 0.9417 & 2.14 \tabularnewline 
No Augmentation & 0.8872 & 3.51 \tabularnewline 
\hline
\end{tabular}}
\end{table*}

The results present high stability in terms of both the Dice coefficient and Hausdorff Distance in the 5-fold cross validation. Moreover, they confirm the importance of data augmentation and maintaining high spatial resolution. Unfortunately, for two cases from the external test set the 95th percentile of HD is significantly higher (for all challenge participants). Since the test data is closed, it is impossible to determine what is the reason, namely whether it is connected to poor algorithm generalizability, the difficulty of particular cases (e.g. aorta dissections), or imperfect annotations.

\begin{table*}[!htb]
\centering
\caption{The comparison of the proposed method to other top-performing SEG.A teams. Only the best submissions from each team are reported for the presentation clarity.}
\renewcommand{\arraystretch}{1.0}
\footnotesize
\resizebox{0.99\textwidth}{!}{%
\begin{tabular}{lccccccc}
\label{tab:results_test}
Team & Case A2C & Case A3C & Case A4C & Case A5C & Case A6C & Mean & Std \tabularnewline
\hline
\multicolumn{8}{c}{Dice Coefficient $\uparrow$} \tabularnewline
\hline
Proposed & 0.907 & 0.907 & 0.912 & 0.923 & 0.935 & 0.917 & \textbf{0.012} \tabularnewline
ATB & 0.885 & 0.891 & 0.918 & 0.917 & 0.934 & 0.909 & 0.020 \tabularnewline
Brightskies & 0.906 & 0.897 & 0.920 & 0.931 & 0.947 & \textbf{0.920} & 0.020 \tabularnewline
NVAUTO & 0.887 & 0.889 & 0.923 & 0.928 & 0.947 & 0.915 & 0.025 \tabularnewline
TeamX & 0.890 & 0.876 & 0.918 & 0.919 & 0.935 & 0.908 & 0.024 \tabularnewline
KARINA & 0.895 & 0.895 & 0.919 & 0.928 & 0.944 & 0.916 & 0.021 \tabularnewline
Culrich & 0.889 & 0.878 & 0.924 & 0.926 & 0.941 & 0.912 & 0.027 \tabularnewline
WGD & 0.887 & 0.877 & 0.921 & 0.917 & 0.938 & 0.908 & 0.025 \tabularnewline
Ouradiology & 0.883 & 0.862 & 0.911 & 0.916 & 0.936 & 0.902 & 0.029 \tabularnewline
\hline
\multicolumn{8}{c}{95th percentile of HD [mm] $\downarrow$} \tabularnewline
\hline
Proposed & 13.50 & 6.95 & 2.92 & 2.48 & 1.88 & 5.55 & 4.88 \tabularnewline
ATB & 10.38 & 7.99 & 2.61 & 2.38 & 1.88 & 5.05 & 3.88 \tabularnewline
Brightskies & 13.25 & 11.79 & 2.61 & 2.23 & 1.59 & 6.30 & 5.71 \tabularnewline
NVAUTO & 13.56 & 5.67 & 2.61 & 2.38 & 1.56 & 5.15 & 4.95 \tabularnewline
TeamX & 13.19 & 10.24 & 2.83 & 3.00 & 2.00 & 6.25 & 5.11 \tabularnewline
KARINA & 16.88 & 4.36 & 2.89 & 2.24 & 2.00 & 5.66 & 6.34 \tabularnewline
Culrich & 15.59 & 7.53 & 2.65 & 2.23 & 2.00 & 6.00 & 5.82 \tabularnewline
WGD & 11.70 & 7.09 & 3.01 & 3.42 & 1.88 & 5.42 & 4.01 \tabularnewline
Ouradiology & 8.64 & 6.92 & 2.97 & 2.62 & 1.86 & \textbf{4.61} & \textbf{2.99} \tabularnewline
\hline
\end{tabular}}
\end{table*}

\subsection{Surface Meshing}

\begin{figure*}[!htb]
	\centering
    \includegraphics[scale=0.58]{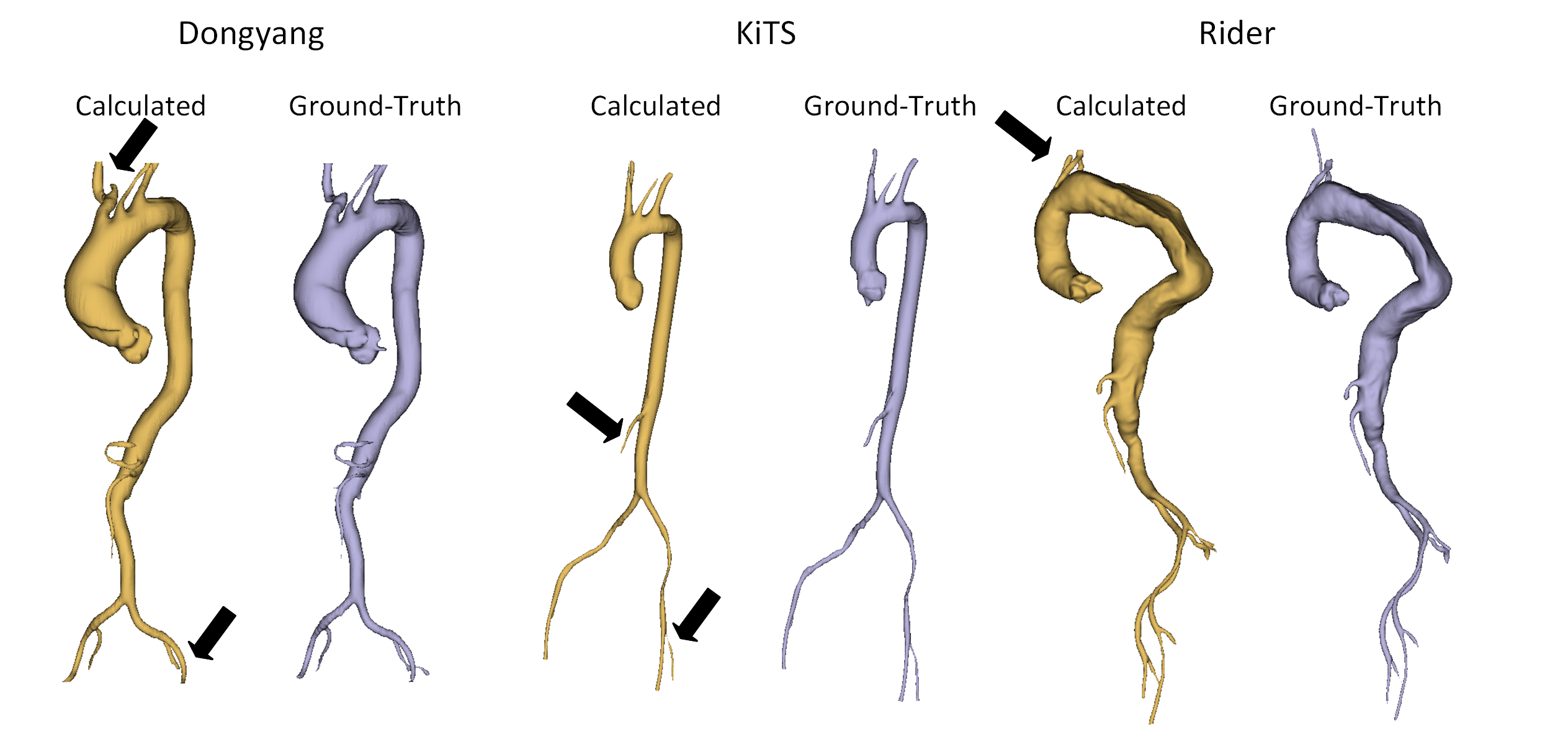}
    \caption{Exemplary surface meshes for cases from all data sources (validation cases). Please note that the differences are mostly in fine-details related to small branches.}   
    \label{fig:surface_mesh}
\end{figure*}

\begin{figure*}[!htb]
	\centering
    \includegraphics[scale=0.7]{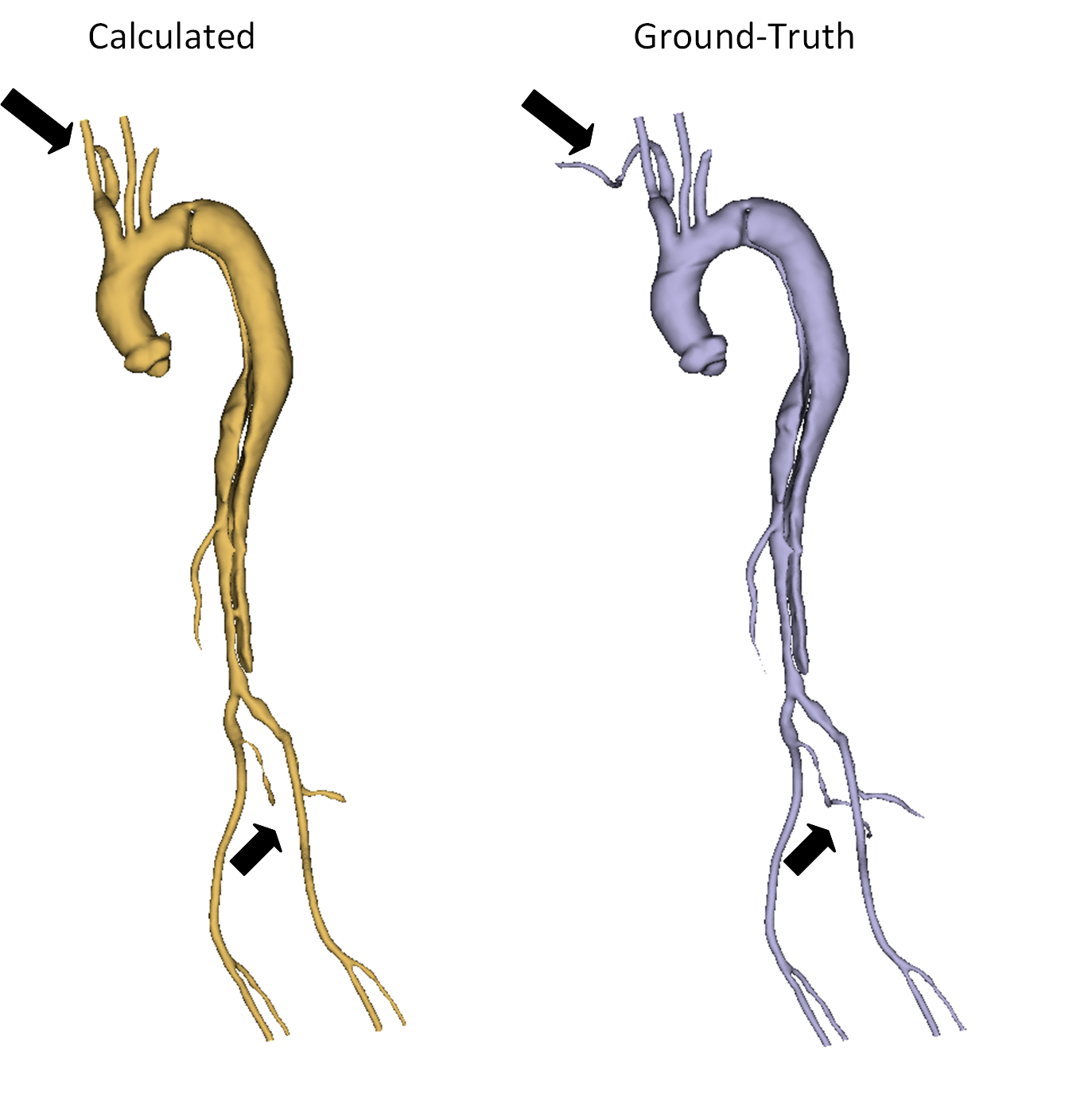}
    \caption{Exemplary surface meshes for a case with an aortic dissection (Rider dataset - validation case). Note that the dissection was correctly segmented, the only differences are related to small branches.}   
    \label{fig:surface_mesh_ad}
\end{figure*}

Exemplary visualizations of the calculated surface meshes are presented in Figure~\ref{fig:surface_mesh}, next to the ground-truth meshes. It can be noted that the calculated meshes are of relatively high quality and the largest disagreements are within the small branches (which is expected). Figure~\ref{fig:surface_mesh_ad} presents visualization for a case with exemplary aortic dissection for which the proposed method generalizes correctly.

The surface meshes were qualitatively evaluated by 8 clinical experts from 5 different hospitals located in Austria, Germany, Iran, and the United Kingdom. The experts evaluated two cases, one with the best quantitative scores, and one with the worst quantitative scores. They were asked to evaluate two aspects, the correctness of the model and the absence of artifacts. The results are presented in Table~\ref{tab:surface_mesh}.

\begin{table*}[!htb]
\centering
\caption{Final results for the clinical evaluation based on the surface meshing. Correctness denotes the ranking in the amount of information that the calculated model correctly communicates to specialists while the absence ranks the resistance to segmentation artifacts.}
\renewcommand{\arraystretch}{1.0}
\footnotesize
\resizebox{0.7\textwidth}{!}{%
\begin{tabular}{lccccc}
\label{tab:surface_mesh}
Team & \multicolumn{1}{c}{Correctness [rank] $\downarrow$} & \multicolumn{1}{c}{Absence [rank] $\downarrow$} & \multicolumn{2}{c}{Final Ranking $\downarrow$} 
\tabularnewline
\hline
Proposed & \multicolumn{1}{c}{1} & \multicolumn{1}{c}{1} & \multicolumn{2}{c}{1} \tabularnewline
NVAUTO & \multicolumn{1}{c}{1} & \multicolumn{1}{c}{2} & \multicolumn{2}{c}{2} \tabularnewline
Brightskies & \multicolumn{1}{c}{3} & \multicolumn{1}{c}{3} & \multicolumn{2}{c}{3} \tabularnewline
Biomed & \multicolumn{1}{c}{4} & \multicolumn{1}{c}{4} & \multicolumn{2}{c}{4} \tabularnewline
Cian & \multicolumn{1}{c}{5} & \multicolumn{1}{c}{5} & \multicolumn{2}{c}{5} \tabularnewline
\hline
\end{tabular}}
\end{table*}

\subsection{Volumetric Meshing}

The quantitative results for the volumetric meshing are presented in Table~\ref{tab:volume_mesh}. Unfortunately, due to the limitations of the Grand Challenge platform, the evaluation was based on the uploaded surface meshes, without the control of the volumetric meshing itself. The differences between participants in this subtask are rather minor and it is hard to conclude which method was the most appropriate, especially since the mesh correctness in terms of medical credibility was not evaluated, only the Jacobian-related statistics.

\begin{table*}[!htb]
\centering
\caption{The final results of the volumetric meshing task (Jac - Jacobian).}
\renewcommand{\arraystretch}{1.0}
\footnotesize
\resizebox{0.99\textwidth}{!}{%
\begin{tabular}{lccccc}
\label{tab:volume_mesh}
Team & Avg. Inv. Elements $\downarrow$ & Median Jac $\uparrow$ & Jac Variance $\downarrow$ & Jac Skewness $\downarrow$ & Final Ranking $\downarrow$
\tabularnewline
\hline
Biomed & 3.49 & 0.570 & $0.72 * 10^{-5}$ & 0.2446 & 1 \tabularnewline
NVAUTO & 3.85 & 0.557 & $0.79 * 10^{-5}$ & 0.2567 & 2 \tabularnewline
Proposed & 4.55 & 0.550 & $2.68 * 10^{-5}$ & 0.8667 & 3 \tabularnewline
Brightskies & 5.98 & 0.545 & $1.69 * 10^{-5}$ & 0.3324 & 4 \tabularnewline
Cian & 61.76 & -100 & $>10^{3}$ & 0.3651 & 5 \tabularnewline
\hline
\end{tabular}}
\end{table*}

\section{Discussion}

The overall outcomes of the proposed method are satisfactory. The method is the most stable in terms of the DC standard deviation among all SEG.A participants. The method achieved DC above 0.9 for all test cases and HD95 below 3mm for three of them. The two cases with significantly larger HD95 were problematic for all challenge participants. Interestingly, it seems that the proposed solution outperformed the nnUNet-based contributions.

There are several limitations of the proposed method. Even though our initial decision about processing the whole 3-D CT volume directly has several advantages, using the patch-based pipeline probably would improve the HD and allow us to perform better segmentation of the small branches. Even though the majority of the structures are clearly visible in the 400x400x400 resolution, the fixed size is not resistant to acquiring different regions of interest that may happen in clinical practice, thus forcing the user to perform manual cropping. Moreover, for the smallest branches, it could be beneficial to upsample the input volumes to resolution far beyond 400x400x400 and use patches with an appropriate shape. We expect that this change could significantly reduce the 95th percentile of HD.

Another limitation of the study itself is the size of the external test set. Unfortunately, just five testing cases make it relatively hard to compare our solution to other challenge participants since it seems that the differences between 7 top-performing teams are statistically insignificant. For all participating teams the median Dice and HD95 was determined by a single test case. Therefore, it is hard to draw any conclusions based on the quantitative results.

Nevertheless, the proposed method has several strong aspects. First, it provides a stable Dice score above 0.9 for all five test cases. With GPU support (which was turned off for the Grand Challenge contribution due to the platform limitation) the inference takes on average less than 2 seconds using NVIDIA A100 GPU and less than one minute without GPU support. It is a significant difference compared to other participants constantly suffering from the time limit strictly enforced in the GD platform (10 minutes per case). Moreover, the proposed method is resistant to different intensity and geometry variations, providing a useful tool in clinical practice in various medical centers.

There are several ideas on how to further improve the method. The first one is to transfer to a patch-based pipeline or to further extend the 3-D volume resolution (e.g. by employing libraries like DeepSpeed~\cite{deepspeed} to reduce the GPU memory consumption). Such modification could improve the segmentation of small branches that are crucial for medical credibility. Another idea to is perform the re-annotation of the ground-truth to separate aorta branches. Even though it would be a time-consuming process, we expect that it would result in a great improvement in small branches segmentation. Moreover, it would be interesting to check the influence of augmentation on methods outside the podium to measure the stability and generalizability, without influencing the final ranking. Finally, the volumetric meshing could be potentially improved by using novel deep learning-based solutions like FlexiCubes~\cite{flexicubes}. However, submission using a different volumetric meshing algorithm would require improvements to the Grand Challenge platform.

To conclude, we proposed a stable and robust method for automatic aorta segmentation. The method achieved competitive results in the SEG.A challenge and after minor improvements could potentially provide even better results. The source code is publicly available, as well as the model used for the final submission.

\subsubsection{Acknowledgements} We gratefully acknowledge Polish HPC infrastructure PLGrid support within computational grant no. PLG/2023/016239.
%
%
\bibliographystyle{splncs04}
\bibliography{bibliography}
\end{document}